\documentclass[10pt,twocolumn,letterpaper]{article}
\usepackage{graphicx} 
\usepackage[table]{xcolor} 
\usepackage{array} 
\usepackage{cvpr} 
\usepackage{multirow}

\definecolor{cvprblue}{rgb}{0.21,0.49,0.74}
\usepackage[pagebackref,breaklinks,colorlinks,allcolors=cvprblue]{hyperref}

\def\paperID{1814} 
\def\confName{CVPR}
\def\confYear{2026}

\title{RGBX-R1: Visual Modality Chain-of-Thought Guided Reinforcement Learning for Multimodal Grounding}

\author{Jiahe Wu$^{1,2,3}$ \quad
Bing Cao$^{1}$ \quad
Qilong Wang$^{1}$ \quad
Qinghua Hu$^{1}$ \quad
Dongdong Li$^{4}$ \quad
Pengfei Zhu$^{1,2,3}$\\
\\
$^{1}$ School of Artificial Intelligence, Tianjin University, Tianjin, China\\
$^{2}$ Low-Altitude Intelligence Lab, Xiong'an National Innovation Center, Xiong'an, China\\
$^{3}$ Xiong'an Guochuang Lantian Technology Co., Ltd., Xiong'an, China\\
$^{4}$ College of Electronic Science and Technology, National University of Defense Technology, Changsha, China\\
\vspace{-13pt}
}

\begin{document}
\maketitle

\begin{abstract}
Multimodal Large Language Models (MLLM) are primarily pre-trained on the RGB modality, thereby limiting their performance on other modalities, such as infrared, depth, and event data, which are crucial for complex scenarios. To address this, we propose RGBX-R1, a framework to enhance MLLM's perception and reasoning capacities across various X visual modalities. Specifically, we employ an \textit{\textbf{U}nderstand}–\textit{\textbf{A}ssociate}–\textit{\textbf{V}alidate} (UAV) prompting strategy to construct the Visual Modality Chain-of-Thought (VM-CoT), which aims to expand the MLLMs' RGB understanding capability into X modalities. To progressively enhance reasoning capabilities, we introduce a two-stage training paradigm: Cold-Start Supervised Fine-Tuning (CS-SFT) and Spatio-Temporal Reinforcement Fine-Tuning (ST-RFT). CS-SFT supervises the reasoning process with the guidance of VM-CoT, equipping the MLLM with fundamental modality cognition. Building upon GRPO, ST-RFT employs a Modality-understanding Spatio-Temporal (MuST) reward to reinforce modality reasoning. Notably, we construct the first RGBX-Grounding benchmark, and extensive experiments verify our superiority in multimodal understanding and spatial perception, outperforming baselines by 22.71\% on three RGBX grounding tasks.

\end{abstract}

\vspace{-3pt}
\section{Introduction}
\label{sec:intro}
Recently, multimodal large language models (MLLM) ~\citep{li2024llava, team2024gemini} demonstrate impressive reasoning capabilities across a wide range of tasks. DeepSeek-R1 ~\citep{guo2025deepseek} proves that reinforcement learning can further enhance the  MLLMs' abilities. Subsequent studies ~\citep{shen2025vlm-r1, li2025videochat-r1,dang2025reinforcing} validate the applicability of reinforcement learning in multimodal scenarios.

However, the effectiveness of reinforcement learning fundamentally depends on the intrinsic capability of MLLMs. 
Most existing MLLMs are primarily pre-trained on the RGB modality, which limits their ability to generalize beyond RGB-based scenarios.
Under degraded conditions—such as nighttime, motion blur, or strong illumination—RGB often becomes unreliable. 
In contrast, alternative visual modalities, including thermal infrared, depth, and event cameras (collectively referred to as X modalities), exhibit superior robustness under such challenging conditions.
Nevertheless, current MLLMs lack the capacity to perceive and reason over these non-RGB modalities, hindering the exploitation of complementary multimodal information for visual understanding in complex environments.

\begin{figure}[t]
\centering
\includegraphics[width=1\columnwidth]{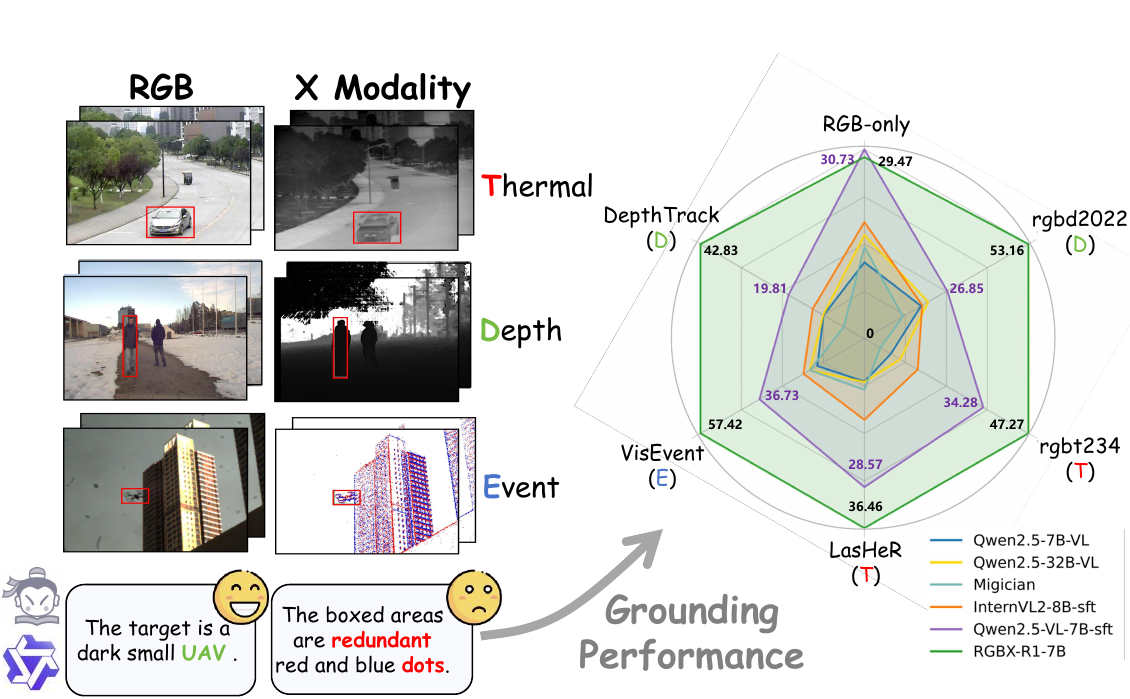} 
\caption{Existing MLLMs exhibit limited capabilities with non-RGB visual modalities and struggle to leverage modality complementarity to enhance multimodal grounding.}
\label{fig1}
\vspace{-5pt}
\end{figure}

To address these challenges, we propose RGBX-R1, a framework that enhances MLLMs' grounding and reasoning capabilities across diverse visual modalities. RGBX-R1 consists of a cold-start data construction and a two-stage training paradigm. Since MLLMs typically lack prior knowledge of non-RGB modalities, they struggle to explore valid RGBX reasoning paths and cannot directly benefit from reward-based reinforcement learning. Therefore, it is necessary to first construct cold-start data serving as strong supervision. 
Specifically, we select RGBX video tracking datasets as the visual source for conducting the cold-start data, as they naturally provide temporally continuous and spatially aligned multimodal data. 
Following Migician~\cite{li2025migician}, we convert RGBX videos into a multi-image grounding dataset tailored for MLLMs, named \textit{RGBX-Grounding}.
Employing RGBX-Grounding as an evaluation benchmark, as shown in Figure~\ref{fig1}, it is revealed that even when MLLMs are supervised fine-tuned with bounding boxes, they still struggle to perceive and understand the X modalities effectively.
For each RGBX-Grounding sample, we construct a corresponding Visual Modality Chain-of-Thought (VM-CoT) to serve as explicit reasoning annotations within the cold-start data. 
We design a multi-step prompting strategy—\textit{\textbf{U}nderstand–\textbf{A}ssociate–\textbf{V}alidate} (UAV) —to transition the MLLM’s RGB cognition to RGBX modality analysis.
Employing Qwen2.5-VL-32B~\cite{bai2025qwen2} as the generator, VM-CoT is produced through the following steps:
1) \textbf{Understand} the target based on the RGB modality;
2) \textbf{Associate} the X modality with RGB semantics via spatial correspondence;
3) \textbf{Validate} the target location by analyzing cross-modal complementarity. 

Based on VM-CoT, we perform the two-stage training paradigm. In the first stage, we employ VM-CoT as supervision to guide MLLM's structured reasoning and establish cognition of the X modality.  
We incorporate a Modality-specific Token Weighting (MTW) mechanism, enabling the model to prioritize reasoning steps most relevant to the current modality. 
This stage equips the model with foundational reasoning capabilities and focuses on modeling the intermediate inference process.
In the second stage, we extend Group Relative Policy Optimization (GRPO)~\cite{shao2024deepseekmath} by introducing a Modality-understanding Spatio-Temporal (MuST) reward mechanism, which comprises three reward components: modality-understanding reward, spatio-temporal reward, and format reward. 
These rewards jointly encourage the model to comprehend modality-specific cues and accurately ground targets within image sequences. 
By sampling multiple responses and optimizing the model based on relative advantage, we guide it toward more accurate reasoning behaviors.

We evaluate mainstream MLLMs on the RGBX-Grounding benchmark. Experimental results show that existing MLLMs struggle to understand X modalities and perform poorly on sequential grounding tasks. In contrast, RGBX-R1, empowered by VM-CoT supervision and the two-stage training paradigm, effectively overcomes these limitations.
RGBX-R1 outperforms supervised fine-tuned baselines by 22.71\% on three RGBX grounding tasks.
It is worth noting that RGBX-R1 demonstrates a phenomenon of ``\textit{Modal Knowledge Emergent}", requiring only a small number of samples to transfer RGB understanding capability to unknown X modalities. 

Our contributions can be summarized as follows:

\begin{itemize}
    \item[$\bullet$] We propose RGBX-R1, the first multimodal visual reinforcement fine-tuning framework that extends the RGB-based cognition of MLLMs to X modalities, empowering MLLMs with robust multimodal grounding.
   
    \item[$\bullet$] We construct RGBX-Grounding, the first dataset that integrates the generated VM-CoT through the 
     UAV prompting strategy, providing high-quality prior supervision for modality understanding and reasoning.
    
    \item[$\bullet$] We introduce a two-stage training paradigm, CS-SFT and ST-RFT, which employs the MTW mechanism and MuST reward to reinforce reasoning capability. Experiments verify our superiority against the competing MLLMs.

\end{itemize}

\section{Related Work}
\subsection{Multimodal Grounding}
Recent advances in multimodal grounding primarily rely on natural language queries to localize targets in visual scenes\citep{deng2023transvg++, su2023language, ji2024progressive}. However, language alone provides limited information compared to the rich cues in high-dimensional visual modalities. To address this, the multi-image grounding (MIG) \citep{bai2025univg, xu2024mc} task emerges, using both language instructions and a template image to locate objects across sequences of images. Nevertheless, these methods remain restricted to the RGB modality, making them vulnerable in challenging scenarios where RGB degrades.

In parallel, RGBX tracking~\citep{hu2025sttrack, hong2024onetracker, cao2024bi, zhu2023vipt} shows that integrating complementary visual modalities—infrared, depth, and event—enhances spatial perception under challenging conditions. Recent efforts \citep{chen2025sutrack, tan2024xtrack, feng2025cstrack, wu2024untrack} further attempt to jointly learn multiple visual modalities through a unified training process.
However, these methods remain confined to the visual domain, lacking explicit decision rationales: even when the target is absent or modality signals fail, they still output a location.
They overlook the potential of language in bridging visual modalities.
In contrast, our work leverages language to explicitly associate RGB and X modalities, enabling interpretable reasoning that captures both modality complementarity and spatial perception.

\begin{figure*}[!t]
\centering
\includegraphics[width=0.98\textwidth]{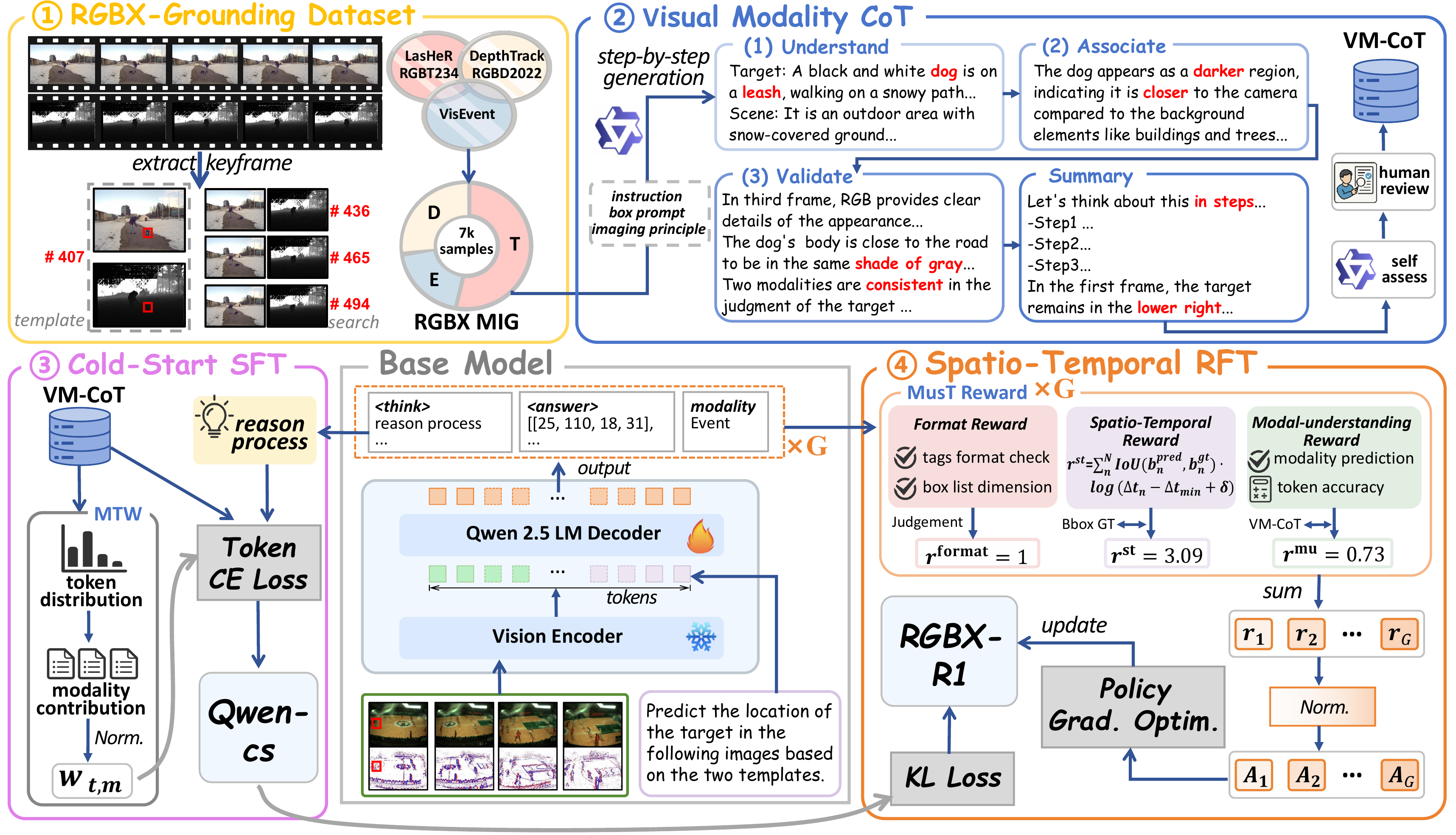}
\caption{
Overall framework of RGBX-R1. (a) RGBX MIG data construction. (b) Constructing VM-CoT with the UAV prompting strategy and two-stage filtering. (c) Cold-start Supervised Fine-Tuning equipped with the MTW mechanism, this stage aims to guide the model in performing structured reasoning. (d) Spatio-Temporal Reinforcement Fine-Tuning. The policy model generates multiple responses and optimizes the model via the MuST group rewards.
}
\label{fig2}
\end{figure*}

\subsection{Reinforcement Learning in MLLMs}
LLMs \citep{grattafiori2024llama, floridi2020gpt, flant5} exhibit strong reasoning capabilities across various tasks. Beyond generating answers, they produce interpretable reasoning steps in natural language through the CoT paradigm \citep{wei2022chain}. These capabilities allow LLMs to decompose complex problems and explain their decision-making steps. In the multimodal domain, MLLMs extend this reasoning ability by integrating visual inputs, evolving into multimodal CoTs~\citep{Mc-cot, Llava-cot, zhang2023multimodalcot}.

Recently, DeepSeek-R1 \citep{guo2025deepseek} adopts the GRPO algorithm, which demonstrates that reinforcement learning~\citep{kumar2024training} can significantly enhance the performance of MLLMs. GRPO eliminates the need for trainable reward models by leveraging rule-based numerical rewards. Subsequent studies further validate the effectiveness of reinforcement learning across multimodal reasoning scenarios, including video reasoning \citep{feng2025video-r1,li2025videochat-r1} and visual grounding \citep{jiang2025rex,bai2025univg}.
Despite these successes, these methods rely heavily on the MLLMs' intrinsic reasoning capacity. However, existing MLLMs lack a fundamental understanding of visual modalities beyond RGB, which constrains their reasoning ability in tasks involving more complex modality information. Moreover, current multimodal CoTs~\citep{wang2025mcot} are unable to jointly accommodate multiple visual modalities, limiting their ability to provide effective guidance for MLLMs in such tasks.

\section{Method}

We propose RGBX-R1, a framework designed to enhance the grounding and reasoning capabilities of MLLMs across diverse visual modalities. As shown in Figure~\ref{fig2}, RGBX-R1 consists of four main components: RGBX-Grounding dataset construction, VM-CoT generation, Cold-start Supervised Fine-Tuning as the first stage, and Spatio-Temporal Reinforcement Fine-Tuning as the second stage.

\subsection{RGBX-Grounding Dataset}
Inspired by Migician's method to process video tracking data, we transform five RGBX tracking datasets~\cite{wang2023visevent, yan2021depthtrack, rgbd2022, li2019rgbt, li2021lasher} into a multi-image grounding (MIG) format.
We sample keyframes at intervals of 24 to 29 frames, and every four keyframes produce eight RGBX images, which are grouped into a single MIG sample. 
In total, we construct 7k sample groups, forming the  \textbf{RGBX-Grounding} dataset. More details of the dataset are in the Appendix.

Formally, we define the RGBX MIG task as follows: given a language query $Q$, an RGB target image $Z_{RGB}$, an X modality target image $Z_{X}$, and $N$ search images $S$ as input, the model $M$ output a list of bounding boxes $B\in\mathbb{R}^{N\times4}$, defined as $B = M(Q, Z_{RGB}, Z_{X}, S)$. In our dataset, $N$ typically is set to 6.

\subsection{Visual Modality CoT}
High-quality supervision is critical for guiding model reasoning. Accordingly, we construct VM-CoT annotations for samples in the RGBX-Grounding dataset, serving as structured reasoning supervision during the cold start stage. We employ Qwen2.5-VL-32B to generate step-by-step reasoning traces under the guidance of predefined prompts and bounding box prompts. The generation follows a three-step pipeline, where the output of each step is incorporated into the prompt for the next step:

\begin{itemize}
    \item[$\bullet$] \textbf{Understand}. Given an RGB template image, the model understands the target’s semantics and the scene context based on the RGB modality.

    \item[$\bullet$]\textbf{Associate}. Provided the paired RGB and X modality template images, along with an explanation of the X modality's imaging principle.
    Leveraging spatial correspondence derived from the bounding box, the model links the target identified in the RGB modality to its counterpart in the X modality.
    
    \item[$\bullet$]\textbf{Validate}. Subsequent search images are analyzed frame by frame to determine whether any modality suffers from information degradation. The model leverages cross-modal complementary information to validate the target’s spatial location.

\end{itemize}

Finally, the outputs from the three reasoning steps are aggregated and fed back into Qwen2.5-VL-32B to produce a structured summary, producing a unified VM-CoT. 
To ensure the quality, we adopt a two-stage filtering process. First, the model self-assesses the plausibility of its reasoning based on the ground-truth bounding boxes. Then, we perform a human review to further remove logically flawed cases. 
While some VM-CoTs may contain errors in details, we believe that they still provide coherent reasoning chains that can guide the model on how to think. In total, we retain more than 5,000 VM-CoTs as high-quality annotations.

\subsection{Cold-Start Supervised Fine-Tuning}
In the first training stage, we perform Cold-Start Supervised Fine-Tuning (CS-SFT) on Qwen2.5-VL with VM-CoTs, enabling it to conduct structured reasoning following our predefined UAV path. The input prompt explicitly specifies the type of X modality, and the VM-CoT serves as the supervision signal under modality-known conditions.

The reasoning process in RGBX grounding typically involves long-form textual generation, and the VM-CoT often contains many generic words (e.g., ``object", ``image") and structural expressions (e.g., ``therefore", ``let me think"), which contribute little to actual modality reasoning. More critically, as different modalities emphasize distinct semantic cues, the contribution of each token to reasoning should vary across modalities.
To address this, we introduce a Modality-specific Token Weighting (MTW) mechanism. 
By measuring the distributional differences of each token across modalities, we estimate its reasoning contribution for the specific modality. 
For a token $t$, we calculate its frequency across all VM-CoTs of the current modality $m$, yielding a normalized distribution $P_m(t)$. We then compute an aggregated mixture distribution $Q_{m^{\prime}}(t)$ over all other modalities $m^{\prime}$.
The modality-specific contribution score of token $t$, denoted as $\mathbf{Contrib}(t,m)$, is quantified via the KL divergence, as Eq.(\ref{eq3}):



\begin{equation} \label{eq3}
\mathbf{Contrib}(t,m)=P_m(t)\cdot\log\left(\frac{P_m(t)}{Q_{m^{\prime}}(t)}\right).
\end{equation}

A higher divergence indicates that the token carries stronger modality-specific information and is more critical in the reasoning process of modality $m$. In CS-SFT, we compute the token-level cross-entropy loss between the generated reasoning process and the target VM-CoT. For each token, its contribution score $\mathbf{Contrib}(t,m)$ is normalized within its modality $m$ to obtain the weight $w_{t,m}$, which is used to reweight the loss:

\begin{equation} \label{eq4}
\mathcal{L}_{\text{sft}} = -\sum_{t=1}^{T} w_{t,m} \cdot \log p_{\theta}\left( y_{t} \mid y_{<t}, x_m \right),
\end{equation}
where $T$ denotes the total number of tokens in the VM-CoT, and $y_t$ represents the target token at the 
$t$-th position.

\subsection{Spatio-Temporal Reinforcement Fine-Tuning}
After the first-stage training, the model, referred to as Qwen-\textit{cs}, acquires foundational modality understanding and step-by-step reasoning along the UAV path. 
In the second stage, we perform Spatio-Temporal Reinforcement Fine-tuning (ST-RFT) on Qwen-\textit{cs}, and adopt the GRPO algorithm to evaluate the quality of responses. Given a query $q$, the policy $\pi_\theta$ samples a group of $G$ candidate responses 
$\mathcal{O} = \{ o_1, o_2, \dots, o_G \}$, and assigns corresponding rewards $\{ r_1, r_2, \dots, r_G \}$ based on predefined rules.
To encourage accurate modality understanding and precise grounding, we propose the MuST reward mechanism, which decomposes the overall reward $r$ into three components: Spatio-Temporal reward $r^\textbf{st}$, Modality-understanding reward $r^\textbf{mu}$ and Format reward $r^\textbf{format}$.

\textbf{Spatio-Temporal Reward.} 
We observe that MLLMs often exhibit a phenomenon we refer to as ``\textit{Inertial Numeric Guessing}” when predicting sequences of bounding boxes. Specifically, the MLLM tends to generate coordinate sequences that deviate from actual visual content, often showing numeric progressions or oscillation patterns around the initial coordinates. 
To mitigate this, we design the Spatio-Temporal Reward $r^\textbf{st}$ by incorporating frame-aware weighting to the IoU-based reward, 
which emphasizes accurate grounding in later frames while reducing the impact of coincidental correctness in earlier ones:

\begin{equation} \label{eq5}
r^{st} = \sum_{n=1}^{N} \log_{10} \left( \Delta t_n - \Delta t_{\min} + \delta \right) \cdot \mathbf{IoU} \left( b^{\text{pred}}_{n}, b^{\text{gt}}_{n} \right),
\end{equation}
where $\Delta t_n$ is the frame interval between the $n$-th image and the template, and $\Delta t_{\min}$ is the minimum frame interval in the sequence. $\mathbf{IoU} $ denotes Intersection over Union metric.

\textbf{Modality-understanding Reward.} 
We incorporate a proportion of modality-unknown samples during training. For these samples, the model is required to additionally perform modality classification. 
If the predicted modality is correct, we compute the reward $r^\textbf{mu}$ based on the mean token accuracy of the reasoning process enclosed in \texttt{<think></think>}, compared against the VM-CoT. For correctly classified cases, $r^\textbf{mu}$ typically ranges between 0.6 and 0.9. If classification is wrong, $r^\textbf{mu} = 0$.

\textbf{Format Reward.}
This reward ensures that the response of the model strictly follows the required format, producing the corresponding content within the delimiters and matching the list dimensions:
``\texttt{<think>}\textit{thinking process}\texttt{</think> <answer>} $\boldsymbol{B}\in\mathbb{R}^{N\times 4}$\texttt{</answer>}''.
If the format is correct, $r^\textbf{format} = 1$. Otherwise, $r^\textbf{format} = 0$.

The final reward for the response $o_i$ is computed as $r_i= r^\textbf{mu} + r^\textbf{st} + r^\textbf{format}$.
Then we compute the group-normalized advantage $\hat{A}_i$ with reward standardization:
\begin{equation} \label{eq6}
    \hat{A}_i = \frac{r_i - \text{mean}(\{r_i\}_{i=1}^G)}{\text{std}(\{r_i\}_{i=1}^G)}.
\end{equation}

The objective of GRPO updates the policy $\pi_\theta$ by maximizing the expected $\hat{A}_i$, encouraging the generation of responses with higher quality. The objective also considers preventing the optimized policy $\pi_\theta$ from deviating far from the Qwen-CoT $\pi_{ref}$ by a KL-divergence term $\mathbb{D}_{\mathrm{KL}}\left(\cdot\|\cdot\right)$:

\begin{equation}\label{eq7}
\begin{aligned}
\texttt{max} \mathbb{E}_{\substack{q,\{o_i\}\sim \pi_\theta^{\text{old}}}}
\left[ \frac{1}{G} \sum_{i=1}^G \frac{1}{|o_i|} \sum_{t=1}^{|o_i|}
\text{clip}(r_{i,t}(\theta), 1 \!\pm\! \epsilon) \hat{A}_{i,t} \right] \\
- \beta \cdot \mathbb{D}_{\mathrm{KL}}\left(\pi_\theta \| \pi_{\text{ref}}\right),
\end{aligned}
\end{equation}

\begin{equation}
    r_{i,t}(\theta) = \frac{\pi_\theta(o_{i,t} \mid q, o_{i,<t})}
                          {\pi_\theta^{\text{old}}(o_{i,t} \mid q, o_{i,<t})},
\end{equation}
where $\beta$ is a hyperparameter of KL regularization towards the reference model.
By leveraging these advantages, the model is optimized toward more accurate reasoning. 

\section{Experiments}
\subsection{Experimental Setup}

\textbf{Dataset.} 
The dataset RGBX-Grounding is constructed from five RGBX tracking datasets as the image sources: VisEvent~\cite{wang2023visevent} for the event modality, DepthTrack~\cite{yan2021depthtrack} and RGBD2022~\cite{rgbd2022} for the depth modality, RGBT234~\cite{li2019rgbt} and LasHeR~\cite{li2021lasher} for the thermal infrared modality. RGBX-Grounding contains a total of 7,432 samples and 59k images. Detailed statistics are provided in the Appendix.

\textbf{Evaluation Metrics.} 
The evaluation on RGBX-Grounding adopts the traditional metric Acc@0.5, which measures grounding accuracy. A prediction is considered correct if its IoU with the ground truth is greater than 0.5. The final score is computed by equally averaging correctness across all frames in a sequence.
Evaluation adopts two settings: modality-known and modality-unknown, based on whether the type of modality is specified in input prompts.

\textbf{Implementation Details.}
RGBX-R1 is developed based on Qwen2.5-VL-Instruct as the base model, and we conduct experiments with both the 3B and 7B versions. In the first stage, we train for 3 epochs on the full RGBX-Grounding dataset, resulting in Qwen-\textit{cs}. In the second stage, we sample from the dataset and designate 20\% of the samples as modality-unknown (with prompts referring to X modality). We then perform 1,500 steps of reinforcement learning, 
obtaining the final RGBX-R1 model. 
For the final version, the hyperparameters are set as $G=8$, $\beta=0.05$, and $\delta=5$. The maximum generation length is set to 2,048.
All experiments are performed on NVIDIA RTX A6000 GPUs.

\textbf{Baselines.}
Our baseline models can be categorized into the following groups:
(1) Basic MLLMs, including LLaVA-OV-7B~\cite{li2024llava-ov}, InternVL2-8B~\cite{cai2024internlm2}, and Qwen2.5-VL~\cite{bai2025qwen2} ranging from 3B to 32B, where the 32B model serves as the generator for VM-CoT. We also incorporated the closed-source model Qwen3-Plus~\cite{yang2025qwen3}.
(2) Models fine-tuned for multi-image grounding tasks, including Migician~\cite{li2025migician} and UniVG-R1~\cite{bai2025univg}. 
(3) Supervised fine-tuned baselines: we fine-tune Qwen2.5-VL-7B and InternVL2-8B on RGBX-Grounding for supervised fine-tuning, denoted as Qwen2.5-VL-7B\textit{-sft} and InternVL2-8B\textit{-sft}. 
We also perform a second-stage training on Qwen2.5-VL-7B\textit{-sft} employing the same ST-RFT as applied to RGBX-R1, denoted as Qwen2.5-VL-7B\textit{-sft+st}.

\subsection{Overall Performance}
\textbf{RGBX Grounding.}
Table \ref{table_2} presents the RGBX grounding performance comparison between our method and the baselines. 
Our method includes the first-stage-only model (\textit{-cs}), second-stage-only model (\textit{-st}) and the two stage model (RGBX-R1). 
To ensure that the baselines can understand the task and input information, we design task-specific prompts for each MLLM and perform evaluation under modality-known conditions.
As shown in Table \ref{table_2}, the grounding performance of several 7B-scale MLLMs remains below 20\%. Even the Qwen-2.5-VL-32B model, which serves as the VM-CoT generator, performs poorly without an explicit reasoning prompt. UniVG-R1~\cite{bai2025univg} fails to comprehend the task and cannot generate answers in sequence format. In contrast, RGBX-R1 achieves substantial improvements across all evaluation subsets. 
After two-stage training, RGBX-R1 outperforms the supervised fine-tuning baselines, with improvements of 15.45\%, 25.43\%, and 23.69\% on the three modality tasks compared to Qwen2.5-VL\textit{-sft}, and 26.37\%, 31.87\%, and 36.11\% compared to InternVL2\textit{-sft}. Even with the 3B model, RGBX-R1 surpasses the 7B\textit{-sft} baselines.

\begin{table}[!t]
  \centering
      \caption{Performance comparison on the RGBX-Grounding, including models ranging from 3B to 32B in scale, and the closed-source model. Two models (\textit{-sft}) serve as the primary baselines, along with Qwen at different training stages as the ablation study.
  }
  {\Huge
  \renewcommand{\arraystretch}{1.07}

  \resizebox{1\linewidth}{!}{
  \begin{tabular}{lcccccc}
    \toprule
\multirow{2}{*}{Method} 
& \multicolumn{2}{c}{Thermal}
& \multicolumn{2}{c}{Depth}
& {Event} & \multirow{2}{*}{AVG.} \\
\cline{2-6}
~ & LasHeR &{RGBT.} & {DepthT.} & {RGBD.} & {VisEvent} & \\
\midrule
{Qwen2.5-VL-32B}~\cite{bai2025qwen2} & 8.33 & 10.34 & 10.76 & 20.70 & 19.13 & {13.24}   \\ 
{Qwen3-Plus}~\cite{yang2025qwen3} & 15.73 & 13.62 & 13.49 & 21.88 & 20.96 & {16.59}   \\ 
\midrule
Qwen2.5-VL-3B~\cite{bai2025qwen2} & 5.87 & 3.59 & 5.05 & 8.59 & 11.22 & 7.70\\
Qwen2.5-VL-3B\textit{-cs} & 9.07 & 11.35 & 11.29 & 12.89 & 22.38 & 14.67\\ 
RGBX-R1-3B & 26.71 & 32.76 & 25.32 & 34.38 & 38.36 & 31.70\\
\midrule
LLaVA-OV-7B~\cite{li2024llava-ov} & 1.78 & 1.47 & 1.58 & 0.78 & 8.44 & 4.22\\
InternVL2-8B~\cite{cai2024internlm2} & 8.67 & 6.47 & 7.38 & 10.16 & 18.85 & 12.24\\ 
Qwen2.5-VL-7B~\cite{bai2025qwen2} & 8.25 & 7.61 & 10.65 & 18.36 & 16.51 & 11.90\\
Migician-~\cite{li2025migician} & 9.84 & 4.60 & 5.38 & 12.50 & 19.07 & 12.47\\ 
\midrule
InternVL2-8B\textit{-sft} & 15.61 & 15.37 & 17.83 & 18.75 & 22.36 & 17.60\\
Qwen2.5-VL-7B\textit{-sft} & 28.57 & 34.28 & 19.81 & 26.56 & 36.73 & 31.04\\  
\midrule
Qwen2.5-VL-7B\textit{-cs} & 14.66 & 17.67 & 19.30 & 23.04 & 25.08 & 19.64\\
Qwen2.5-VL-7B\textit{-st} & 25.14 & 30.15 & 14.55 & 16.87 & 22.74 & 23.19\\
Qwen2.5-VL-7B\textit{-sft+st} & 29.75 & 39.86 & 22.24 & 24.05 & 38.98 & 32.74\\ 
RGBX-R1-7B & 36.46 & 47.27 & 42.83 & 53.16 & 57.42 & 46.53\\
\bottomrule
  \end{tabular}}}
  \label{table_2}
\end{table}

\textbf{Single Modality Grounding.}
To assess the discrimination capability of different MLLMs for the RGB and X modalities, we further evaluate the 7B-scale models using a single modality. We adopt modality-unknown prompts and provide either RGB or X images as input. As shown in Table \ref{table_3}, ``RGB" denotes using only RGB inputs across all subsets, while ``X-total" denotes using only the corresponding X modality.
When relying solely on RGB, RGBX-R1 outperforms Qwen2.5-VL-7B by 17.74\%, attributed to the Spatio-Temporal reward improving sequential coordinate prediction. RGBX-R1 lags behind the baseline Qwen2.5-VL-7B\textit{-sft} by 1.96\%.
When relying solely on the X modality, RGBX-R1 surpasses Qwen2.5-VL-7B\textit{-sft} by 15.83\%, 11.63\%, and 8.58\% across the three modalities, and exceeds the pretrained model by 17.01\% in total. In contrast, Qwen2.5-VL-7B\textit{-sft} improves over the pretrained model by 16.91\% when using only RGB, but by merely 4.54\% when using only X. These results indicate that training solely with coordinate supervision provides limited capability for effective perception of the unknown X modality.

\subsection{Discussion}
\textbf{Training Stage.}
Table \ref{table_2} presents the impact of different training stages on the grounding performance. RGBX-R1 achieves an average improvement of 26.89\% compared to the first-stage-only Qwen2.5-VL-7B\textit{-cs}, demonstrating the complementary benefits of the two-stage training pipeline. When using only the second stage, the performance of Qwen2.5-VL-7B\textit{-st} is even lower than Qwen2.5-VL-7B\textit{-sft} on the Depth and Event modalities and exhibits a 23.40\% overall performance drop compared to the two-stage RGBX-R1. This gap underscores the necessity of the cold start with the VM-CoT. 
Compared to the baseline Qwen2.5-VL-7B\textit{-sft+st} that combines conventional SFT and ST-RFT, RGBX-R1 has achieved comprehensive improvements, with an average increase of 13.79\%. This improvement is attributed to the supervision of the reasoning process by VM-CoT.

\begin{table}[!t]
\centering
\renewcommand{\arraystretch}{1.0}
\caption{Performance comparison of 7B-scale models under unimodal inputs with unknown types.}
\resizebox{0.97\linewidth}{!}{ 
\begin{tabular}{cccccc}
\toprule
Method & RGB & Thermal & Depth & Event & X-total\\
\midrule
Qwen2.5-VL & 12.82 & 5.67 & 6.13 & 10.44 & 7.54\\
Migician & 14.90 & 6.23 & 4.89 & 11.86 & 8.19\\
InternVL2\textit{-sft} & 18.93 & 9.64 & 7.17 & 11.86 & 10.12\\
Qwen2.5-VL\textit{-sft} & 32.48 & 12.30 & 7.44 & 13.57 & 12.08\\ 
\midrule
RGBX-R1 & 30.56 & 28.13 & 16.08 & 22.15 & 24.55\\
\bottomrule
\end{tabular}
}
\label{table_3}
\end{table}

\begin{table}[!t]
\centering
\renewcommand{\arraystretch}{1.15}
\caption{Ablation study of the MTW mechanism on RGBX-R1-7B across the two training stages.}
\resizebox{1\linewidth}{!}{ 
\begin{tabular}{ccccc|cc}
\toprule
\multirow{2}{*}{Version} & \multirow{2}{*}{MTW}  & \multicolumn{3}{c|}{Grounding} & \multicolumn{2}{c}{Rensoning} \\
\cline{3-5} \cline{6-7}
 &  & {Th.} & {De.} & Ev. & TA. & Mu. \\
\midrule
Qwen2.5-VL\textit{-cs} & & 13.64 & 17.08 & 22.74 & 64.15 & - \\
Qwen2.5-VL\textit{-cs} & \checkmark & 13.82 & 18.07 & 23.08 & 62.39 & - \\
RGBX-R1 &  & 36.95 & 36.73 & 55.65 & - &  75.51\\
RGBX-R1 & \checkmark & 39.31 & 42.05 & 57.06 & - & 79.62 \\
\bottomrule
\end{tabular}
}

\label{table_6}
\end{table}

\textbf{MTW Mechanism.}
We conduct an ablation study on the MTW mechanism to evaluate its impact on both grounding and reasoning performance across four model variants, as shown in Table~\ref{table_6}. 
To measure reasoning performance, we adopt the mean token accuracy (TA.) as the metric for the first-stage model (\textit{-cs}), and adopt the mean Modality-understanding reward (Mu.) over the last 50 steps for two-stage models.
As shown in Table \ref{table_6}, the MTW mechanism consistently improves grounding performance in both stages, with a more pronounced gain in the complete two-stage model. Moreover, incorporating MTW in the first stage does not affect the training stability. Even when evaluated by mean token accuracy, the model with MTW shows only a drop of 1.76\%. In practice, its reasoning ability becomes more modality-specialized, which translates into stronger grounding capability.

\begin{figure}[t]
\centering
\includegraphics[width=0.96\columnwidth]{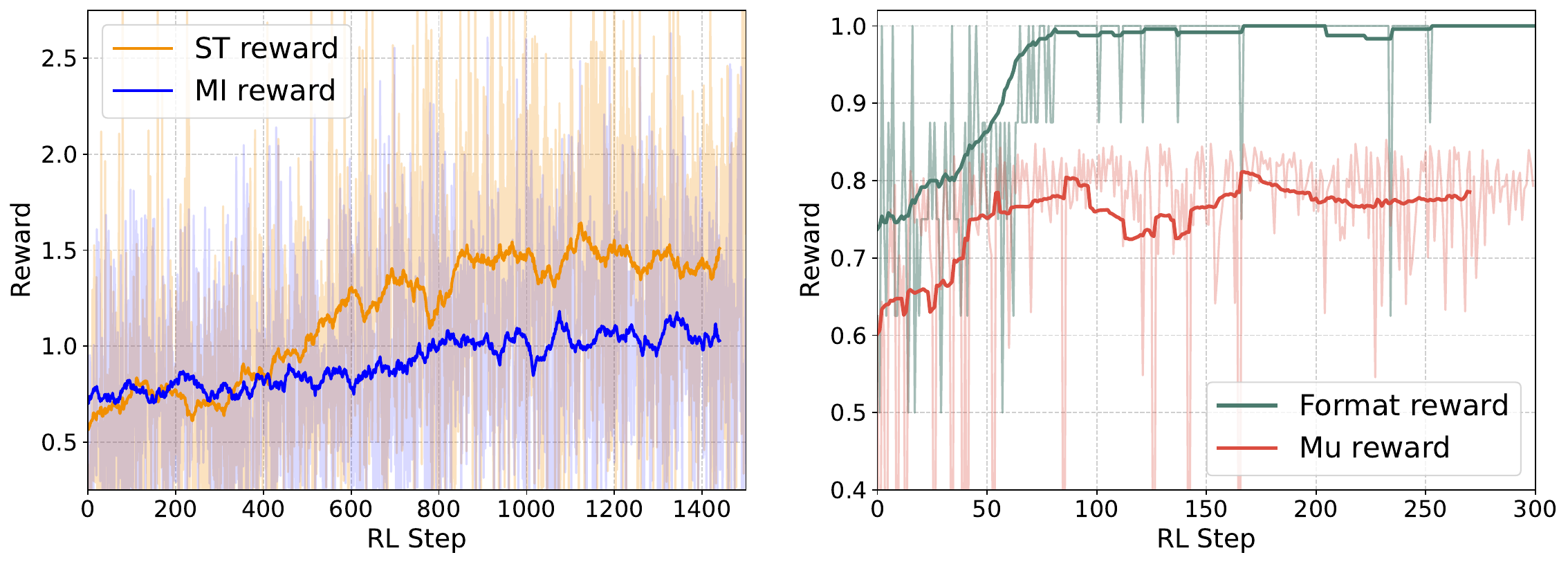} 
\caption{The curves of rewards during ST-RFT.
}
\label{fig4}
\end{figure}

\textbf{MuST Reward.}
The ablation studies on the reward mechanism under modality-unknown evaluation are presented in Table~\ref{table_4}. \textbf{Case 4} represents the complete MuST reward mechanism. \textbf{Case 1} demonstrates that the Format reward, serving as the fundamental reward, largely determines the overall effectiveness of training. The performance on the event modalities in Case 1 is even inferior to Qwen2.5-VL-7B\textit{-cs}. 
In \textbf{Case 2}, the depth modality is most affected without $r^\textbf{mu}$, showing a 7.42\% drop compared to Case 4. In \textbf{Case 3}, we replaced our Spatio-Temporal (ST) reward $r^\textbf{st}$ with the Mean IoU (MI) reward $r^\textbf{mi}$, which computes the mean IoU value of all predicted bounding boxes as the reward. ST reward leads to an average 2.81\% improvement in grounding accuracy.
Furthermore, we visualize the evolution of the MuST reward throughout reinforcement steps, applying smoothing to the curves in Figure~\ref{fig4}.
The results confirm that the ST reward facilitates more precise grounding than the MI reward. The Modality-understanding (Mu) reward and Format reward converge within around 200 steps, and the overall training stabilizes around 1,500 steps.

\begin{table}[!t]
\renewcommand{\arraystretch}{1.0}
\centering
\caption{
Ablation study of reward mechanisms on the grounding performance of RGBX-R1-7B. 
}
\resizebox{1.0\linewidth}{!}{
\begin{tabular}{c |c c c c|c c c c}
\toprule
 \makebox[0.02\textwidth][c]{}
 &\makebox[0.04\textwidth][c]{$r^\textbf{format}$} &\makebox[0.025\textwidth][c]{$r^\textbf{mu}$} &\makebox[0.025\textwidth][c]{$r^\textbf{st}$} &\makebox[0.025\textwidth][c]{$r^\textbf{mi}$} &\makebox[0.03\textwidth][c]{Thermal} &\makebox[0.03\textwidth][c]{Depth} &\makebox[0.03\textwidth][c]{Event} &\makebox[0.03\textwidth][c]{Total}
 \\
\midrule
Case 1& & \checkmark& \checkmark& & 26.81 & 25.24 & 23.70 & 25.41\\
Case 2&\checkmark & & \checkmark& & 37.89 & 34.63 & 52.18 & 42.83\\
Case 3&\checkmark & \checkmark &  & \checkmark& 35.43 & 41.68 & 54.67 & 43.62\\
Case 4&\checkmark & \checkmark & \checkmark & & 39.31 & 42.05 & 57.06 & 46.43\\
\bottomrule
\end{tabular}}
\label{table_4}
\end{table}

\begin{table}[!t]
\centering
\caption{Ablation study of the training data scale.}
\renewcommand{\arraystretch}{1.05}
\resizebox{1\linewidth}{!}{ 
\begin{tabular}{cccc|ccc}
\hline
Sample & \multicolumn{3}{c}{Modality-known} & \multicolumn{3}{c}{Modality-unknown}\\
\cline{2-4} \cline{5-7}
Size & Thermal & Depth & Event & Thermal & Depth & Event\\
\hline
base & 8.13 & 12.28 & 16.51 & 8.11 & 10.76 & 15.60\\
1\% & 24.87 & 30.37 & 28.59 & 24.43 & 28.58 & 27.18\\
10\% & 33.67 & 43.21 & 44.12 & 33.57 & 37.09 & 42.36\\
Full & 38.17 & 45.90 & 57.42 & 39.31 & 42.05 & 57.06\\ 
\hline
\end{tabular}
}
\label{table_5}
\end{table}

\begin{figure}[t]
\centering
\includegraphics[width=1\columnwidth]{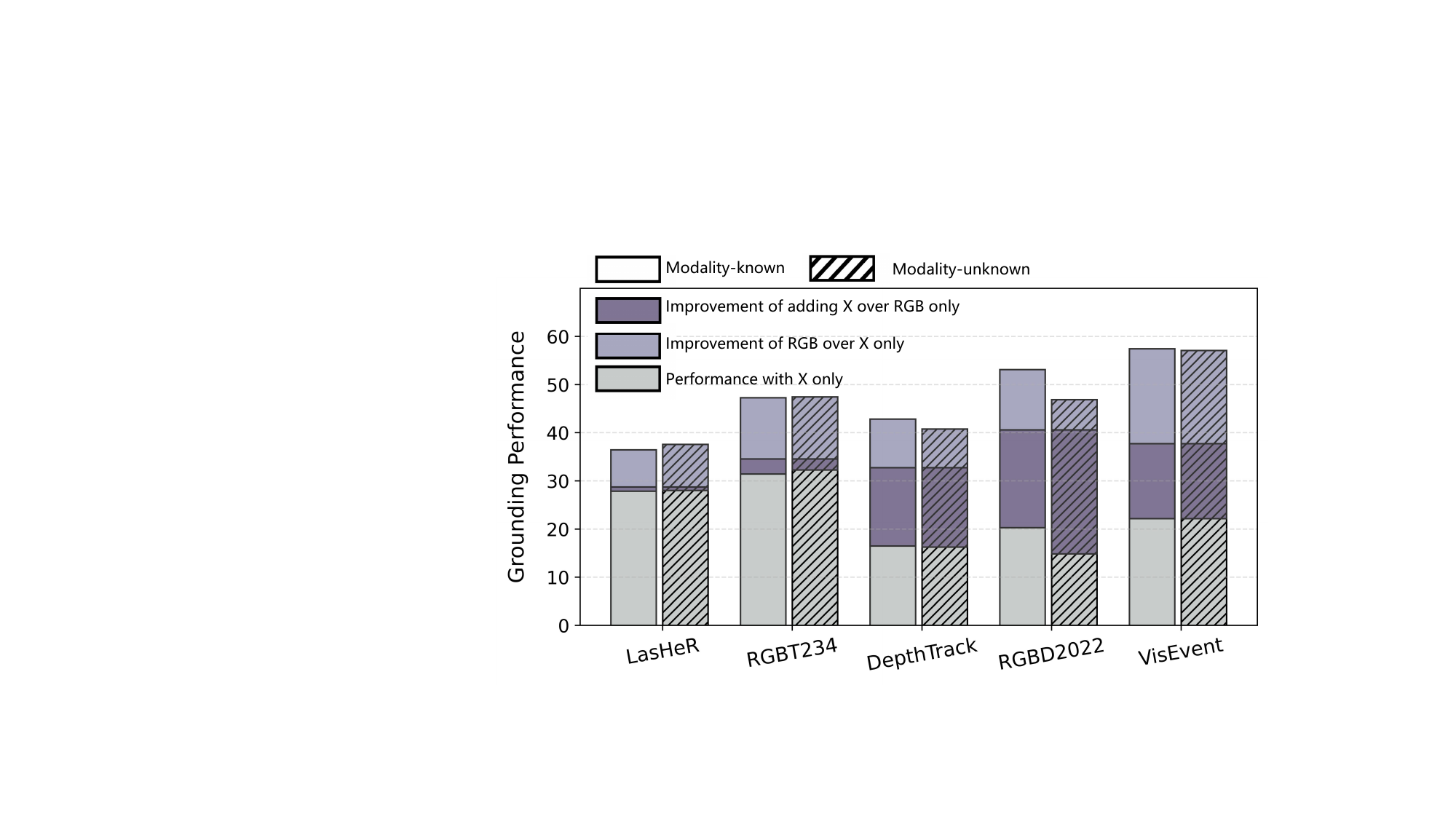} 
\caption{Grounding performance of RGBX-R1-7B under different modality inputs to illustrate modality contributions.
}
\label{fig3}
\end{figure}

\textbf{Modality Contribution.}
In Figure~\ref{fig3}, we illustrate the impact of the RGB and X modalities on grounding performance. The dashed lines represent evaluations under modality-unknown conditions. Across all five subsets, using solely RGB consistently outperforms using only the X modality, with this advantage highlighted in deep purple. The light purple segments highlight the superiority of joint RGB and X modality, where the model leverages complementary information from both modalities to achieve more robust grounding than using either modality alone. Interestingly, in both thermal infrared datasets, modality-unknown inference slightly exceeds modality-known inference. We attribute this to richer cross-modal reasoning triggered by the absence of explicit types of X modalities.

\begin{figure}[!t]
\centering
\includegraphics[width=1.0\columnwidth]{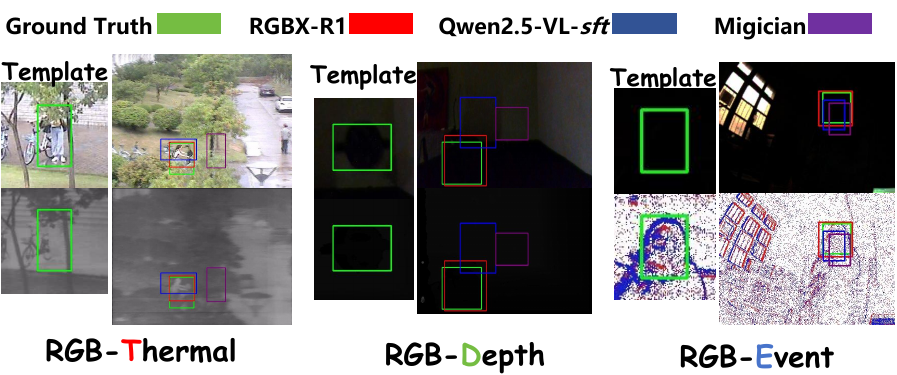} 
\caption{Visualization of RGBX grounding results. The green rectangles indicate targets in the template images.
}
\label{fig5}
\end{figure}

\textbf{Sample Scale.}
We present the effect of training data scale on RGBX-R1’s grounding performance in Table~\ref{table_5}.
We adopt uniform sampling when utilizing only 1\% and 10\% of the total samples. At the 1\% level, each modality includes an average of merely 20 multi-image combinations. Remarkably, even with such a small fraction of data, RGBX-R1 achieves a significant enhancement over the baseline Qwen2.5-VL-7B. With 10\% of the data, the performance on the depth modality already approximates that obtained with the full dataset. 
These results underscore that our method, even with a minimal number of RGBX samples, can effectively elicit the model's understanding of previously unknown X modalities and conduct modal complementary analysis. Such performance exhibits a strong property of ``\textit{Modal Knowledge Emergent}".

\begin{figure*}[t]
\centering
\includegraphics[width=0.99\textwidth]{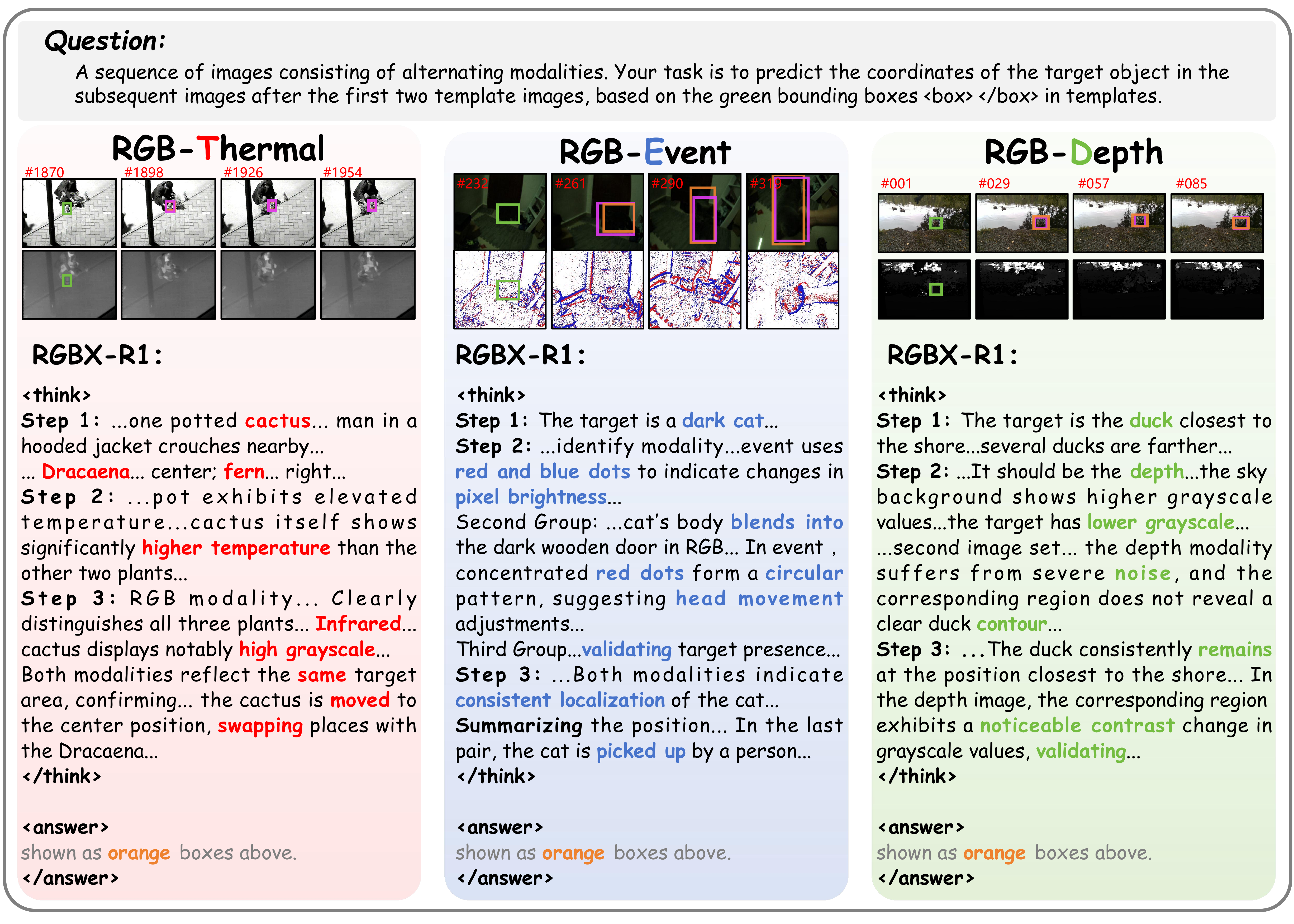}
\caption{
An example of RGBX-R1-7B’s reasoning outputs, with reasoning keywords highlighted in bold and specific color. In the images, green, purple and orange boxes denote the input information, the ground truth and the prediction, respectively.
}
\label{fig6}
\end{figure*}

\subsection{Qualitative Evaluation}
\textbf{Grounding Result.}
We visualize the qualitative results of our method and other approaches, as shown in Figure~\ref{fig5}. All methods are evaluated using 7B scale models. 
While competing methods struggle under degraded conditions, RGBX-R1 consistently maintains robust grounding performance, benefiting from its enhanced understanding of the X modality and cross-modal complementary reasoning. 
In the RGB-Event case, where the RGB modality completely fails, other methods can only estimate the target's location based on its initial position in the template images. Migician exhibits the ``\textit{Inertial Numeric Guessing}” phenomenon in all three cases, deviating from the actual visual content and solely relying on the initial coordinates for prediction. In the RGB-Depth case, even when both modalities suffer from degradation, our method still successfully grounds the target. 
Furthermore, our RGBX-R1 also has a more precise grounding capability, framing the entire object rather than partial regions.

\textbf{Reasoning Process.}
In Figure~\ref{fig6}, we present three representative reasoning examples of RGBX-R1-7B across diverse modalities, showcasing its structured multi-step reasoning process. The model consistently follows the intended reasoning path, beginning with comprehensive target understanding, proceeding through fine-grained frame-by-frame modality analysis, and culminating in a synthesized estimation of the target’s spatial location. Throughout this process, RGBX-R1 actively leverages modality-specific characteristics for both supplementation and cross-checking—illustrated by reasoning segments such as “validating target presence” and “indicating consistent localization.” This explicit integration of cross-modal cues enables the model to refine its spatial perception and significantly bolster grounding robustness. Additional visualization results and reasoning examples are provided in the Appendix.

\section{Conclusion}

In this paper, we present RGBX-R1, a framework that empowers MLLMs with enhanced grounding and reasoning capabilities across diverse visual modalities.
RGBX-R1 first leverages the VM-CoT to provide structured reasoning supervision for Cold-Start Supervised Fine-Tuning, equipping the model with fundamental modality understanding with the aid of the MTW mechanism.
Building upon GRPO, Spatio-Temporal Reinforcement Fine-Tuning employs the MuST reward to reinforce spatio-temporal perception and modality reasoning.
Extensive experiments on our proposed RGBX-Grounding benchmark demonstrate substantial improvements over strong baselines, confirming the effectiveness of our approach in complex and degraded multimodal grounding scenarios. To the best of our knowledge, RGBX-R1 is the first multimodal visual reinforcement fine-tuning framework to explore MLLMs' reasoning capabilities for RGBX grounding tasks, establishing a new task paradigm and benchmark for future work.



{
    \small
    \bibliographystyle{ieeenat_fullname}
    \bibliography{main}
}

\clearpage
\setcounter{page}{1}
\maketitlesupplementary

\setcounter{section}{0}  
\section{RGBX-Grounding Dataset}
RGBX-Grounding covers three modality-specific tasks, sampling image data from five RGBX datasets: VisEvent for the event modality, DepthTrack and RGBD2022 for the depth modality, and RGBT234 and LasHeR for the thermal infrared modality.
To extract keyframes, we select specific frames at an interval of 24 to 29 frames in most cases, depending on the total frame count of each original video. 
However, a small portion of videos in VisEvent, LasHeR, and RGBT234 contain fewer than 80 frames. For these short videos, we adopt a unified minimum interval corresponding to 13 frames. Such short videos account for less than 1\% of the dataset. 
In RGBX-Grounding, 97.24\% of the samples consist of two template images followed by six search images for grounding. Additionally, 2.28\% and 0.48\% of the samples solely contain six and four images, respectively.
To maintain a balanced distribution of data samples, we extract at most 8 sample groups of multi-image grounding from each video, avoiding excessive samples from videos exceeding 2,000 frames.
For training and testing splits, we follow the original splits of VisEvent, DepthTrack, and LasHeR to partition our multi-image samples. Since RGBD2022 and RGBT234 do not provide predefined splits, we perform our own partitioning. In total, RGBX-Grounding comprises 7,432 multi-image grounding samples, with more than 58k images. Detailed statistics are provided in Table \ref{table_1}.

\begin{table}[!h]
\centering
\renewcommand{\arraystretch}{1.2}
\caption{Statistics of the RGBX-Grounding dataset.}
\resizebox{1\linewidth}{!}{ 
\begin{tabular}{c|c|c|c|c|c}
\hline
 & \multicolumn{2}{c|}{Thermal} & \multicolumn{2}{c|}{Depth} & Event \\
\hline
Image Numbers& \multicolumn{2}{c|}{33,082} & \multicolumn{2}{c|}{11,216} & 14,636 \\
\cline{2-3} \cline{4-5} \cline{6-6}
Subset & {LasHeR} & {RGBT234} & {DepthTrack} & {RGBD2022} & {VisEvent} \\
Train Samples & 2,615 & 522 & 732 & 366 & 1,096 \\
Test Samples & 826 & 175 & 240 & 65 & 795 \\
\hline
\end{tabular}
}
\label{table_1}
\end{table}

\section{Performance of RGBX Tracking Methods}
We evaluate RGBX video object tracking methods~\cite{zhu2023vipt,wu2024untrack,chen2025sutrack,sttrack,hou2024sdstrack} on our proposed RGBX-Grounding benchmark. SUTrack and Un-Track are selected as representative methods. We use their official checkpoints, models trained on the original RGBX video datasets, to perform our multi-image grounding task. Their grounding performance results are presented in Table \ref{table_4}, marked with *.
It should be noted that the training samples from only three original RGBX video datasets—LasHeR, DepthTrack, and VisEvent—amount to 1,894k, which is 32 times larger than the RGBX-Grounding dataset. Moreover, their training samples are temporally continuous.
The results indicate that the frame-sampled data from RGBX-Grounding presents a more challenging grounding task compared to the original continuous data, exhibiting a performance degradation of 20\%-30\% compared to the evaluations ~\cite{wu2024untrack, chen2025sutrack} conducted on the original video dataset.
This is primarily because, in the multi-image grounding paradigm, targets may undergo substantial displacement between frames, which poses a greater challenge to a model's capabilities in target identification and relational reasoning.

\begin{table}[!t]
\centering
\renewcommand{\arraystretch}{1.18}
\caption{Grounding performance of RGBX video tracking methods on RGBX-Grounding.}
\resizebox{1\linewidth}{!}{ 
\begin{tabular}{ccccc}
\hline
& RGBT234 & RGBD2022 &  VisEvent  & AVG.\\
\hline
\rowcolor{gray!15} 
\multicolumn{5}{c}{Trained on original video data} \\ 
Un-Track*~\cite{wu2024untrack} & 60.58 & 61.18 & 51.65 & 53.76 \\
SUTrack\_b224*~\cite{chen2025sutrack} & 64.16 & 62.70 & 53.87 & 56.16 \\
\hline
\rowcolor{gray!15} 
\multicolumn{5}{c}{Trained on RGBX-Grounding} \\ 
ViPT~\cite{zhu2023vipt} & 46.34 & 39.36 & 46.11 & 45.72\\
Un-Track~\cite{wu2024untrack} & 47.28 & 38.49 & 46.59 & 46.19\\
SDSTrack~\cite{hou2024sdstrack} & 48.05 & 39.78 & 46.61 & 46.42\\
STTrack~\cite{sttrack} & 49.96 & 42.73 & 46.75 & 47.04\\
SUTrack\_t224~\cite{chen2025sutrack} & 49.16 & 40.09 & 46.78 & 46.74\\
SUTrack\_b224~\cite{chen2025sutrack} & 50.44 & 40.47 & 46.92 & 47.11\\
RGBX-R1 & 47.27 & 53.16 & 57.42 & 55.43\\
\hline
\end{tabular}
}
\label{table_4}
\end{table}

\begin{table}[!t]
\centering
\caption{Grounding performance of RGBX video tracking methods on RGBX-Grounding with 1\% data scale.}
\renewcommand{\arraystretch}{1.1}
\resizebox{1\linewidth}{!}{ 
\begin{tabular}{ccccc}
\hline
& RGBT234 & RGBD2022 &  VisEvent & AVG.\\
\hline
ViPT &  19.62 & 17.62 & 12.61 & 14.11\\
Un-Track & 19.47 & 17.90 & 12.84 & 14.27\\
STTrack & 20.37 & 19.26 & 16.97 & 17.69\\
SUTrack\_t224 & 22.13 & 17.89 & 13.99 & 15.61\\
SUTrack\_b224 & 22.46 &  18.37 & 15.28 & 16.69\\
RGBX-R1 & 24.43 & 28.58 & 27.18 & 26.80\\
\hline
\end{tabular}
}
\label{table_5}
\end{table}


For a fair comparison, we retrained these RGBX trackers on the RGBX-Grounding training data using the pre-trained weights of the backbone networks~\cite{itpn, ostrack}. The results are summarized in Table \ref{table_4}. Among these, only SUTrack~\cite{chen2025sutrack} and our RGBX-R1 utilize modality-mixed samples for a single training session, whereas other approaches are trained and evaluated under modality-specific conditions. The results show that our RGBX-R1 achieves the best average performance across the three X-modal grounding tasks, outperforming the second-best SUTrack\_b224 by an average of 8.32\%. 
However, these RGBX trackers generally exhibit better grounding performance in the thermal infrared modality, with SUTrack\_b224 surpassing our RGBX-R1 by 3.17\%. The performance advantage of RGBX-R1 is most pronounced in the depth modality, where it exceeds the suboptimal SUTrack\_b224 by 12.69\%.

Additionally, to evaluate the cross-modal transfer capability of these RGBX trackers under extreme few-shot settings, we retrained them using only 1\% of modality-balanced samples. 
As shown in Table \ref{table_5}, we trained with only 1\% of the training samples and evaluated on the full test set. 
On the three RGBX grounding tasks, our method demonstrates a more significant performance advantage, surpassing the best RGBX tracker, STTrack~\cite{sttrack}, by 4.06\%, 9.32\%, and 10.21\%, respectively.

\section{Prompts in UAV Prompting Strategy}

We annotate reasoning processes for all training samples in RGBX-Grounding. Using Qwen2.5-VL-Instruct-32B as the generator, we employ the UAV prompting strategy to produce the Visual Modality Chain-of-Thought (VM-CoT) by 4 steps in total. The first three steps correspond to the three stages of the UAV strategy, while the final step reformats and summarizes the outputs from these steps to produce the final VM-CoT. In the UAV strategy, each step’s prompt consists of multiple sub-prompts, including \{task prompt\}, \{modality prompt\}, \{previous-step answer prompt\}, and \{current-step prompt\}. 
The \{previous-step answer prompt\} incorporates the generator’s output from the preceding step. 
The prompts for the first three steps follow the sub-prompt structure described below:
\begin{itemize}
    \item[$\bullet$] \textbf{Step-1}. \{task prompt\} +  \{current-step prompt\}.
    \item[$\bullet$] \textbf{Step-2}. \{task prompt\} +  \{modality prompt\} + \{previous-step answer prompt\} + \{current-step prompt\}.
    \item[$\bullet$] \textbf{Step-3}. \{task prompt\} +  \{modality prompt\} + \{previous-step answer prompt\} + \{current-step prompt\}.

\end{itemize}

The specific sub-prompt content is presented below. The \{task prompt\}, which aims to guide the model to understand the input format and the final task objective. The \{\} denotes the specific parameters to be passed. The \{task prompt\} is as follows:

\textit{A total of \{num\} images are provided, consisting of alternating RGB and \{modality\} images. Each RGB image is paired with the subsequent \{modality\} image, which is spatially and temporally aligned. The final task is to predict the target’s location in each of the subsequent images, based on the target marked by the green bounding box. Green box coordinates: \{box\}.}

The \{modality prompt\} provides the imaging principle of the X modality as prior knowledge:
\begin{itemize}
    \item[$\bullet$] \textbf{Thermal: }   
    \textit{The thermal infrared modality is based on temperature sensing. Regions with higher temperatures have higher grayscale values, appearing white or light gray; conversely, lower temperatures appear as dark gray or black. For example, a pedestrian target in the RGB modality exhibits a higher temperature and appears as a gray-white human-shaped region in the infrared modality, while cooler backgrounds such as roads appear black.}
    \item[$\bullet$] \textbf{Depth: }   
    \textit{The depth modality is based on the distance of objects from the camera. The closer an object is to the viewpoint, the lower its grayscale value, appearing dark gray or black; the farther it is, the whiter it appears. For instance, if the target is a ball positioned near the camera, it appears as a dark circular region in the depth modality.}
    \item[$\bullet$] \textbf{Event: }   
    \textit{The event modality is based on changes in pixel intensity. A response is triggered only when brightness increases or decreases beyond a threshold, with red indicating increased intensity and blue indicating decreased intensity. For example, the headlights of a car in the RGB modality appear as small circular clusters of red pixels in the event modality, indicating the emergence of a brighter object in that region.
    }
\end{itemize}

Below are the \{current-step prompt\} used in each step of the UAV strategy, where \{example\} denotes a carefully designed fixed example provided as guidance for each step:
\begin{itemize}
    \item[$\bullet$] \textbf{Understand: }   
    \textit{Using only the RGB modality, describe the target inside the green box, along with any other objects or scene context related to the target (if applicable). For example, \{example\}.}
    \item[$\bullet$] \textbf{Associate: }   
    \textit{Establish spatial correspondence between the two modalities. Based on the RGB modality and the description of the last step, interpret the target information at the corresponding positions in each {modality} image. Here is the principle of \{modality\}: \{modality prompt\}. For example, \{example\}. The reasoning process for this step should not exceed 200 words.}
    \item[$\bullet$] \textbf{Validate: }   
    \textit{First, analyze the complementary relationship between the two modalities in target grounding. Specifically, for each image, determine whether one modality experiences information degradation and whether the other modality provides complementary cues. Avoid general statements about modality contribution—perform image-specific analysis. Finally, verify the target location based on the cross-modal complementarity and describe the target position. For example, \{example\}. The reasoning process for this step should not exceed 400 words.}
\end{itemize}

Finally, we merge the outputs from the above three steps and feed them back into the generator for structured summarization. During this process, redundant content across the three outputs is filtered out, and a unified format for VM-CoT is produced. After a two-stage filtering process, we retain 5,000 VM-CoTs
for the samples in RGBX-Grounding.

\section{Modality-specific Token Weighting}
In the Modality-specific Token Weighting (MTW) mechanism, we measure the distributional differences of each token across modalities.
Inspired by the CIDEr metric~\cite{vedantam2015cider}, we quantify the importance of tokens for modal reasoning through frequency statistics. Similar to the concept of Term Frequency and Inverse Document Frequency in CIDEr, if a word appears more frequently within the current modality, it is considered more discriminative for that specific modality. Conversely, if a token is common across the VM-CoTs of all modalities, its value is deemed lower.
For a token $t$, we statistics its frequency in all VM-CoTs of the specific modality $m$, denoted as $f(t, m)$. 
We then compute its normalized distribution $P_m(t)$ for the current modality $m$, and the aggregated mixture distribution $Q_m(t)$ over all other modalities $m^{\prime}$:
\begin{equation} \label{eq1}
P_m(t)=\frac{f(t,m)}{N_m},
\end{equation}

\begin{equation} \label{eq2}
Q_m(t)=\frac{\sum_{m^{\prime}\neq m}f(t,m^{\prime})}{\sum_{m^{\prime}\neq m}N_{m^{\prime}}},
\end{equation}
where $N_m$ is the total number of tokens of all VM-CoTs in modality $m$. We then quantify the modality-specific contribution score of token $t$:

\begin{equation} \label{eq3}
\mathbf{Contrib}(t,m)=P_m(t)\cdot\log\left(\frac{P_m(t)}{Q_{m^{\prime}}(t)}\right).
\end{equation}

The contribution scores $\mathbf{Contrib}(t,m)$ are normalized separately for each modality. Specifically, we first apply a logarithmic transformation to obtain $L_m(t)$, followed by min–max normalization. Finally, the normalized values are linearly mapped to the range [0.05,1], denoted as $\alpha_{t,m}$:

\begin{equation} \label{eq3}
\alpha_{t,m} = \mu + (1-\mu) \cdot \frac{\log \left( \max \left( \mathbf{Contrib}(t,m), \delta \right) \right) - L_{\min}}{L_{\max} - L_{\min} + \varepsilon},
\end{equation}

where $\mu$ is set to 0.05, and $\delta$ and $\varepsilon$ are very small positive numbers.
$\alpha_{t,m}$ is used as the training loss weights during Cold-Start Supervised Fine-Tuning. The modality-specific weights obtained from MTW are presented in Table \ref{table_2}.
\begin{table}[h]
\centering
\renewcommand{\arraystretch}{1.2}
\caption{Modality-specific token weights for each modality. We present a part of tokens with the highest and lowest weights.}
\resizebox{1\linewidth}{!}{
\setlength{\tabcolsep}{6pt}
\begin{tabular}{cc|cc|cc}
\hline
\multicolumn{2}{c|}{Thermal} & \multicolumn{2}{c|}{Event} & \multicolumn{2}{c}{Depth} \\
\hline
token & weight & token & weight & token & weight \\
\hline
thermal & 1.00  &  event  & 1.00  & depth & 1.00  \\
heat & 0.97  &  dots  & 0.95  & camera & 0.92  \\
infrared & 0.94  &  blue & 0.93  & proximity & 0.90 \\
signature & 0.89  &  points & 0.93  & distance & 0.87 \\
temperature  & 0.88  &  motion  & 0.92  & darker & 0.86  \\
cooler & 0.87  & red  & 0.91  & farther & 0.86 \\
bright  & 0.85  &  drone  & 0.90  & lighter & 0.85  \\
umbrella & 0.85  &  rapid  & 0.88  & background & 0.84  \\
warmer& 0.83  & movement  & 0.86  & closer & 0.83 \\
\vdots & \vdots & \vdots & \vdots & \vdots & \vdots \\
the  &  0.05  &  a & 0.05  & of  & 0.05  \\
\hline
\end{tabular}
}
\label{table_2}
\end{table}

\section{Grounding Visualization}

\begin{figure*}[h]
\centering
\includegraphics[width=0.84\textwidth]{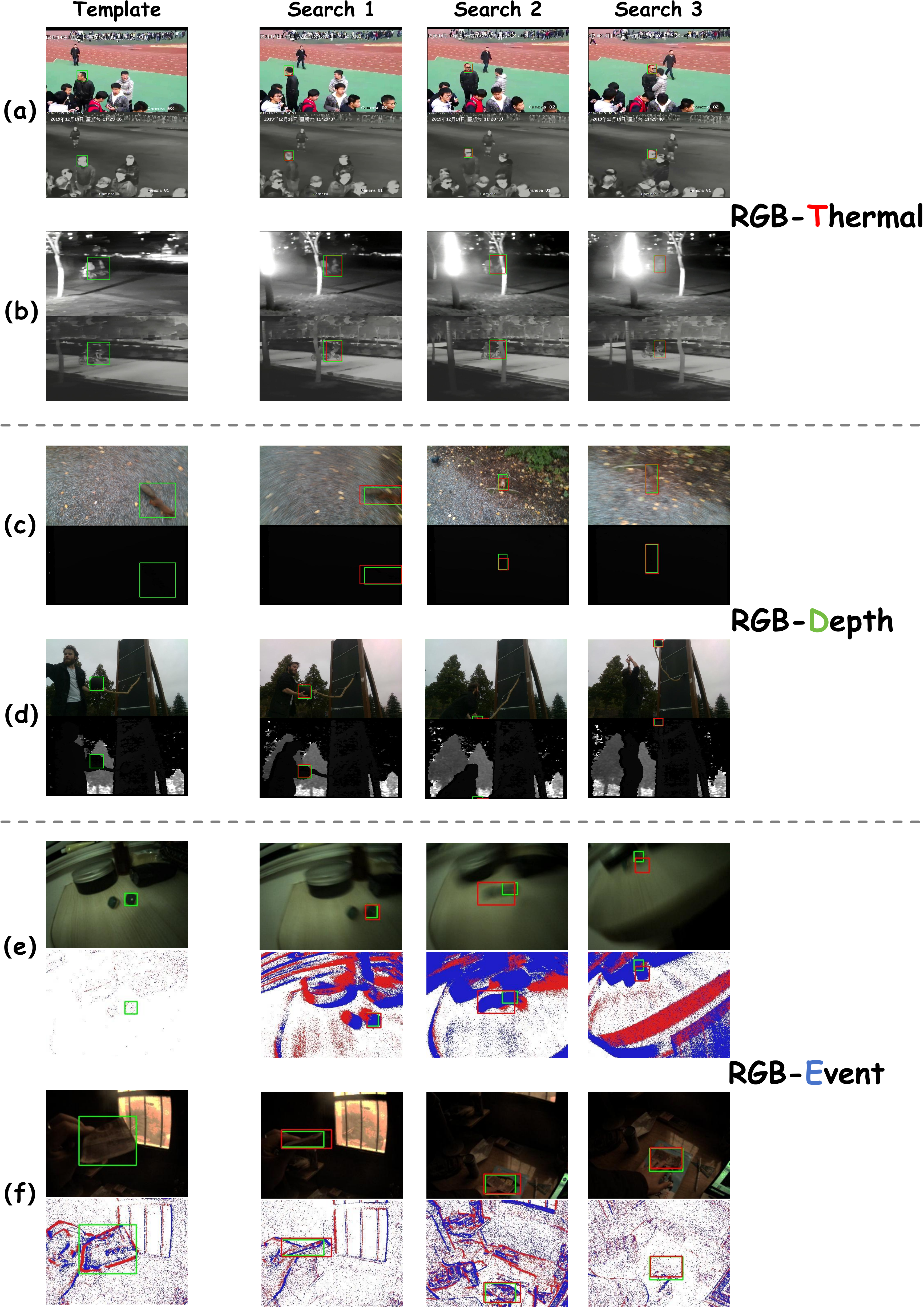}
\caption{
Visualization of RGBX-R1's grounding results. Green boxes denote ground-truth bounding boxes, while red boxes indicate RGBX-R1's predictions.
}
\label{fig7}
\end{figure*}

In Fig.\ref{fig7}, we present the grounding results of RGBX-R1, where two complete multi-image grounding examples are shown for each RGBX modality. Among cases b,d,c,e, various scene degradation phenomena are observed to validate the robustness of our model’s grounding capability.

In case b, the third search image exhibits RGB overexposure, yet the model accurately localizes the target by relying on infrared information.
In case c, the search images display sparse depth information because the target and background are at nearly the same distance, causing partial information loss. The model combines the blurred RGB view and partial depth contours to estimate the target’s position.
In case d, the target has completely moved out of the frame in the second search image. The model infers its approximate disappearance position based on its relative relation to other objects.
In case e, motion blur occurs in the second and third search images of the RGB modality. The model relies on the event modality to determine the approximate target position; however, relying solely on the event modality still results in minor grounding bias.
In addition, in cases a and f, the model localizes the target with high precision.

\section{Reasoning Examples}

\begin{figure*}[!h]
\centering
\includegraphics[width=0.92\textwidth]{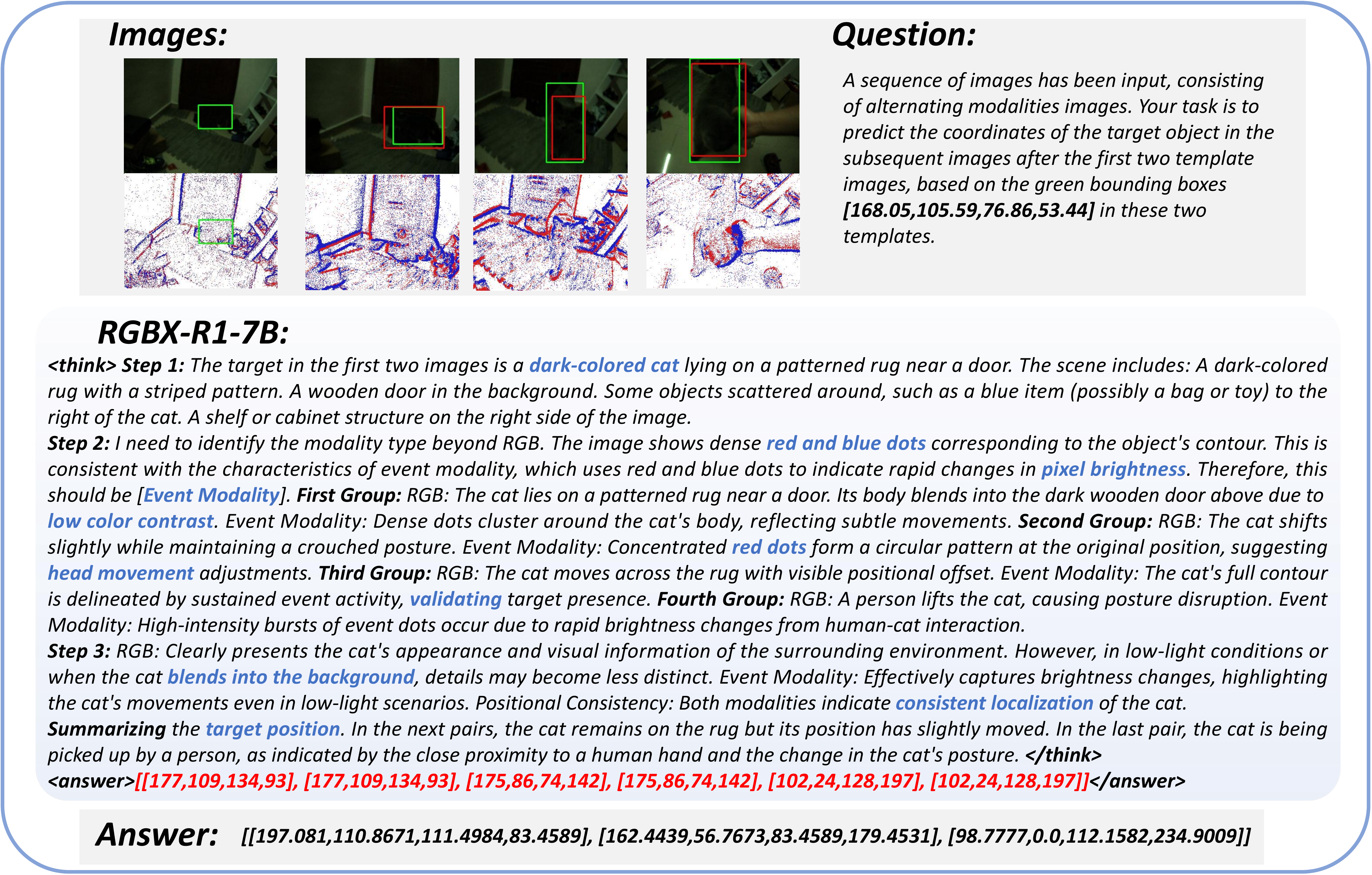}
\caption{
An example reasoning process of correct RGB–Event grounding. The target to be grounded is a cat.
}
\label{fig8}
\end{figure*}

\begin{figure*}[!h]
\centering
\includegraphics[width=0.92\textwidth]{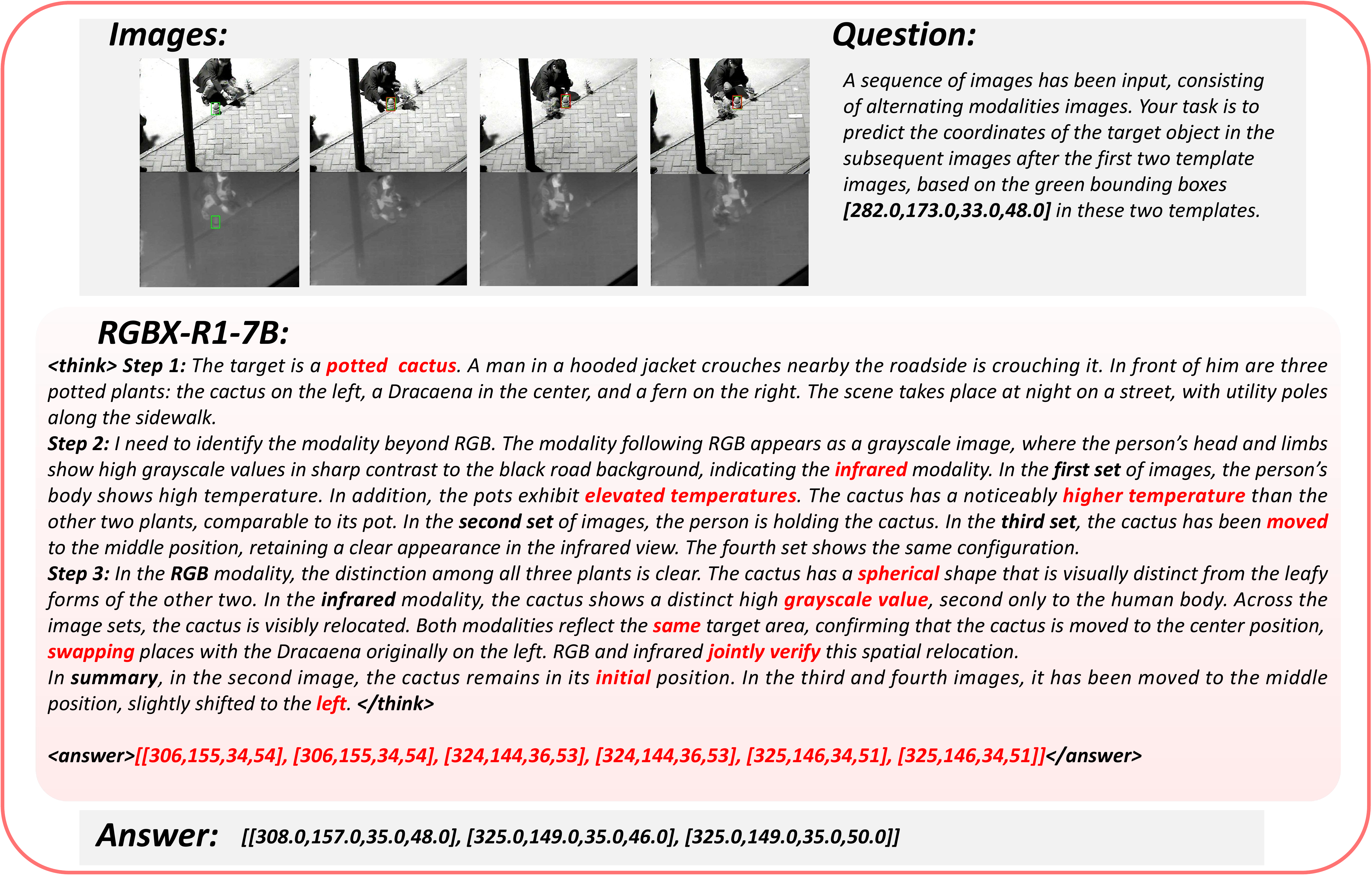}
\caption{
An example reasoning process of correct RGB–Thermal grounding. The target to be grounded is a pot of cactus.
}
\label{fig9}
\end{figure*}

\begin{figure*}[!h]
\centering
\includegraphics[width=0.92\textwidth]{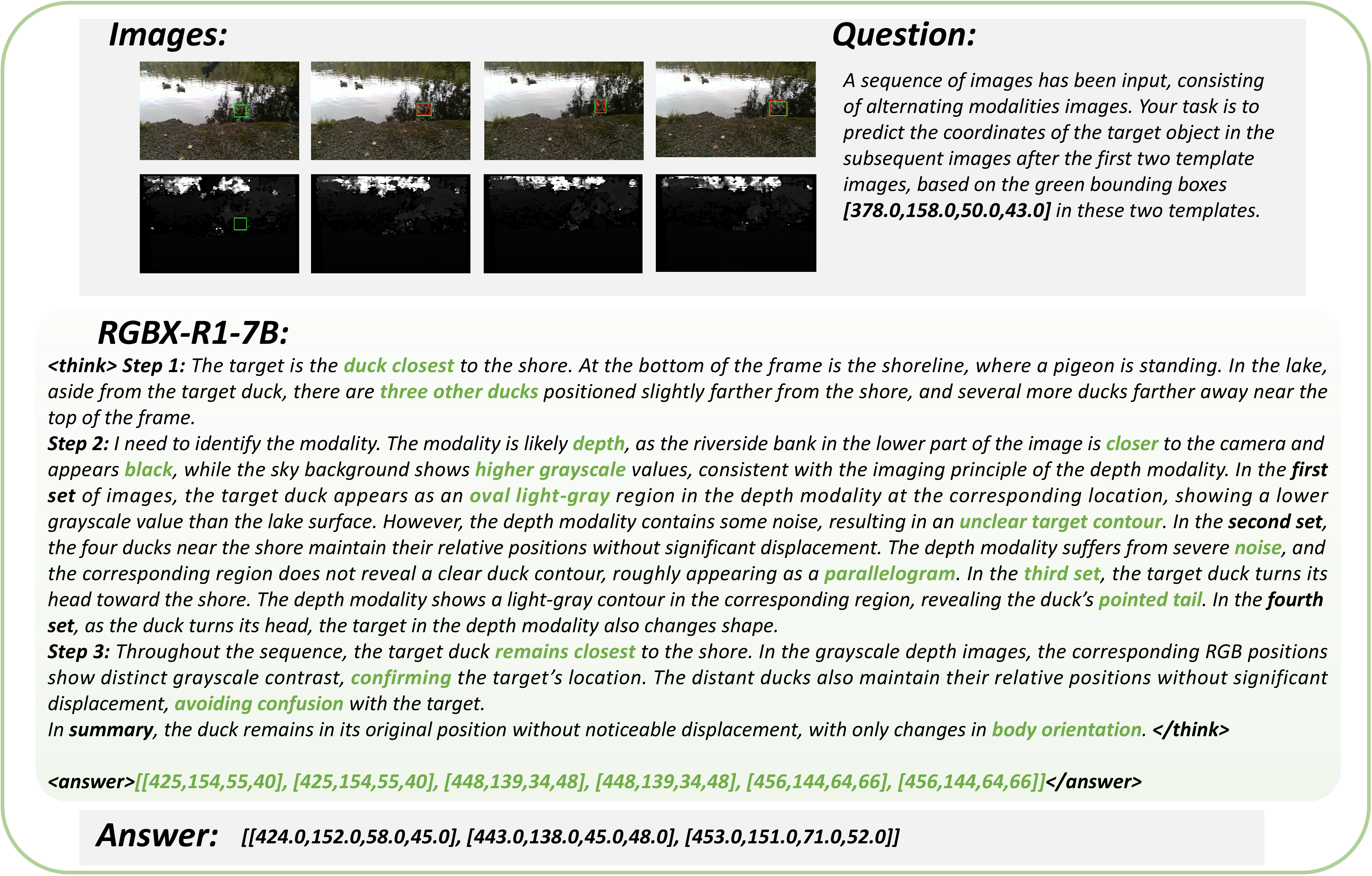}
\caption{
An example reasoning process of correct RGB–Depth grounding. The target to be grounded is a duck.
}
\label{fig10}
\end{figure*}

We present the complete reasoning processes for several representative case studies included in the paper. Fig.\ref{fig8}, Fig.\ref{fig9} and Fig.\ref{fig10} correspond to event, thermal infrared, and depth modalities, respectively. From these reasoning processes, RGBX-R1 strictly follows the predefined reasoning path. Building upon a solid understanding of each modality’s characteristics, it leverages cross-modal complementarity to achieve collaborative grounding.

\begin{figure*}[!h]
\centering
\includegraphics[width=0.85\textwidth]{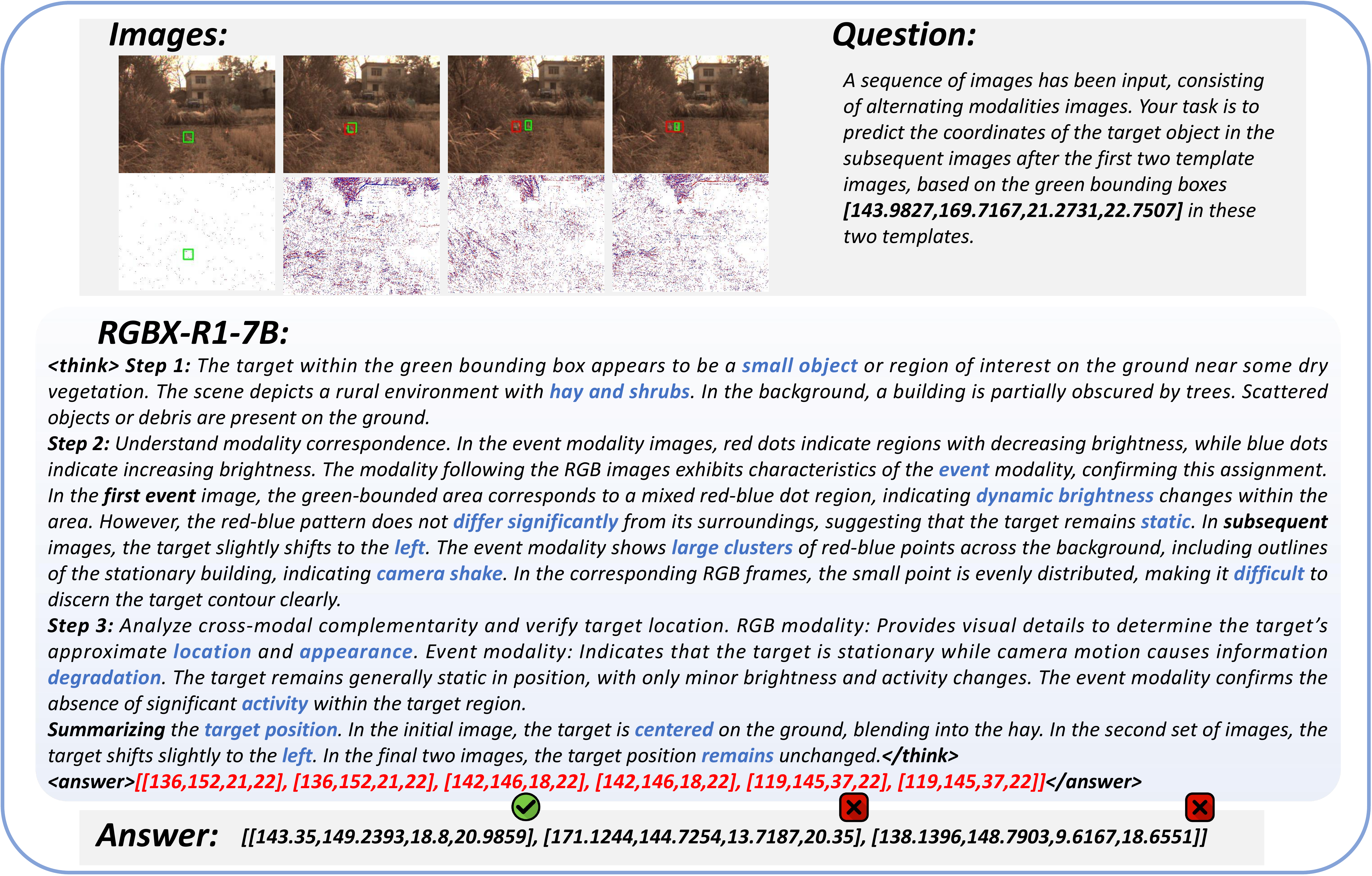}
\caption{
An example reasoning process of RGB–Event grounding with incorrect predictions. The second and third search images contain grounding errors. The target to be grounded is a chicken.
}
\label{fig11}
\end{figure*}

\begin{figure*}[!h]
\centering
\includegraphics[width=0.83\textwidth]{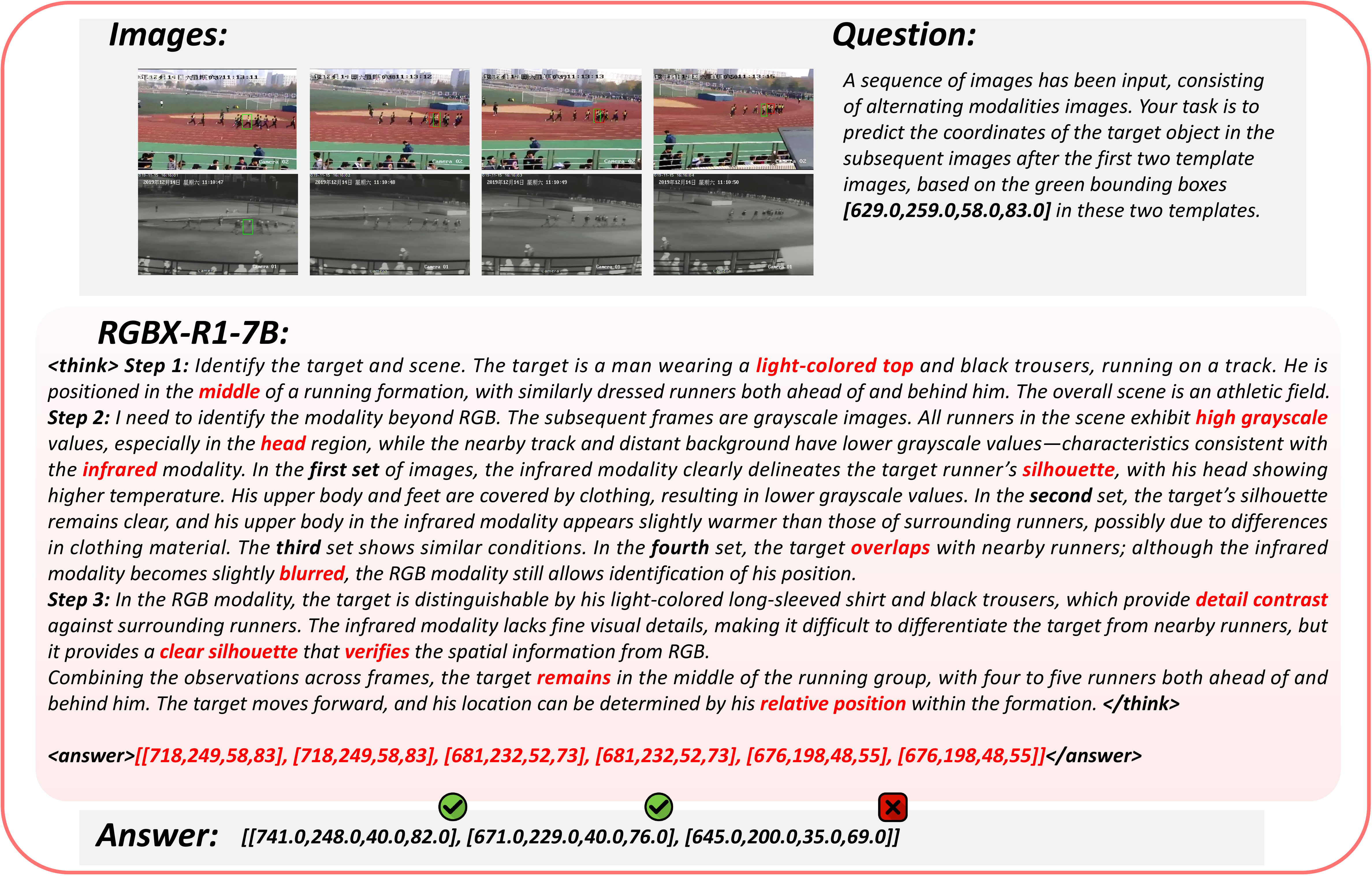}
\caption{
An example reasoning process of RGB–Thermal grounding with incorrect predictions. The third search image contains the grounding error. The target to be grounded is a man.
}
\label{fig12}
\end{figure*}

\begin{figure*}[!h]
\centering
\includegraphics[width=0.83\textwidth]{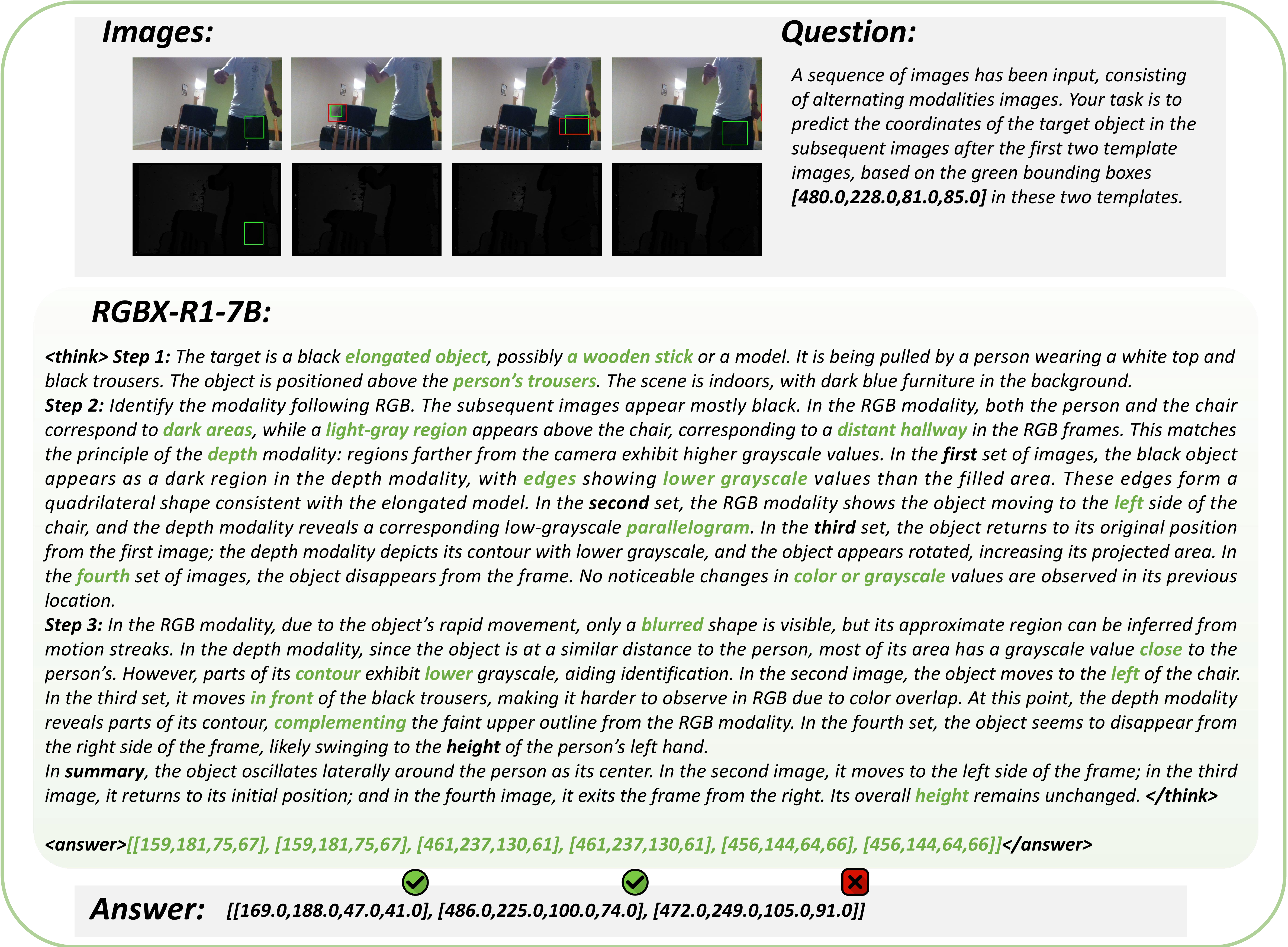}
\caption{
An example reasoning process of RGB–Depth grounding with incorrect predictions. The third search image contains the grounding error. The target to be grounded is a notebook.
}
\label{fig13}
\end{figure*}

In addition to the correct grounding cases with positive reasoning outcomes, we also examine failure cases to analyze their reasoning processes. As shown in Fig.\ref{fig11}, Fig.\ref{fig12} and Fig.\ref{fig13}, each presents a negative case from different modalities. According to the IoU@0.5 metric, correct boxes and incorrect ones are marked with green check marks and red crosses, respectively.
In Fig.\ref{fig11}, due to noise present in both the RGB and event modalities, the model fails to accurately interpret the target semantics.
In Fig.\ref{fig12}, the target overlaps or swaps positions with visually similar objects, leading to incorrect identification.
In Fig.\ref{fig13}, in the final search image, both the color contrast in RGB and grayscale contrast in the depth modality are low, causing the model to mistakenly infer that the target has exited the frame.

\end{document}